\documentclass[11pt]{article}
\usepackage{hyperref}
\usepackage{coling2018}
\usepackage{times}
\usepackage{url}
\usepackage{latexsym}

\usepackage{graphicx}  
\usepackage{bm}
\usepackage{amssymb, amsmath, eqnarray}
\usepackage{float}
\usepackage{subfigure}
\usepackage{booktabs}
\usepackage{threeparttable}
\usepackage{textcomp}
\usepackage{CJKutf8}
\usepackage{xstring}

\newcommand*\myat{{\fontfamily{ptm}\selectfont\small @}}

\title{Refining Source Representations with Relation Networks \\ for Neural Machine Translation}

\author{Wen Zhang \\
  Affiliation / Address line 1 \\
  Affiliation / Address line 2 \\
  Affiliation / Address line 3 \\
  {\tt email@domain} \\\And
  Second Author \\
  Affiliation / Address line 1 \\
  Affiliation / Address line 2 \\
  Affiliation / Address line 3 \\
  {\tt email@domain} \\}
  
\def\first{$^1$}
\def\second{$^2$}
\def\third{$^3$}

\def\comma{$^,$}
\author{Wen Zhang\first\comma\second ~~~ Jiawei Hu\first\comma\second ~~~ Yang Feng\first\comma\second\thanks{\hspace{0.15cm}Corresponding author.} ~~~ Qun Liu\third\comma\first
\\
\\
{ \first {Key Laboratory of Intelligent Information Processing, Institute of Computing Technology, CAS}}   \\
{ \second {University of Chinese Academy of Sciences}}   \\
{\tt \{\href{mailto:zhangwen@ict.ac.cn}{zhangwen},\href{mailto:hujiawei@ict.ac.cn}{hujiawei},\href{mailto:fengyang@ict.ac.cn}{fengyang}\}\myat ict.ac.cn}	\\
{ \third {ADAPT Centre, School of Computing, Dublin City University}}\\
\href{mailto:qun.liu@dcu.ie}{{\tt qun.liu\myat dcu.ie}} \\
}

\date{}

\begin{document}
\begin{CJK*}{UTF8}{gbsn}

\maketitle
\begin{abstract}

	Although neural machine translation with the encoder-decoder framework has achieved great success recently, it still suffers drawbacks of forgetting distant information, which is an inherent disadvantage of recurrent neural network structure, and disregarding relationship between source words during encoding step. Whereas in practice, the former information and relationship are often useful in current step. We target on solving these problems and thus introduce relation networks to learn better representations of the source. The relation networks are able to facilitate memorization capability of recurrent neural network via associating source words with each other, this would also help retain their relationships. Then the source representations and all the relations are fed into the attention component together while decoding, with the main encoder-decoder framework unchanged. Experiments on several datasets show that our method can improve the translation performance significantly over the conventional encoder-decoder model and even outperform the approach involving supervised syntactic knowledge.

\end{abstract}

\section{Introduction}
\label{sec1}

%
%
\blfootnote{
    %
    %
    %
    %
    %
    %
    \hspace{-0.65cm}  
    This work is licensed under a Creative Commons 
    Attribution 4.0 International License.
    License details:
    \url{http://creativecommons.org/licenses/by/4.0/}
}

In recent years, Neural Machine Translation (NMT) \cite{kalchbrenner2013recurrent,sutskever2014sequence,Bahdanau} has achieved great success in some language pairs, rivalling the state-of-the-art Statistical Machine Translation (SMT). The Recurrent Neural Network (RNN) encoder-decoder architecture is widely used framework for NMT, the principle behind which is that: encoding the meaning of the input bidirectionally into a concept space via RNNs and decoding into target words with RNNs based on this encoding \cite{sutskever2014sequence,Bahdanau}. This means that encoding principle leads to a deeper understanding and learning of the translation rules, and hence better translation than conventional SMT that considers only surface forms, e.g., words and phrases.

The RNNs with gating, such as Gated Recurrent Unit (GRU) \cite{cho2014learning} or Long Short-Term Memory (LSTM) \cite{hochreiter1997long}, are designed to memorize useful history information and meanwhile forget irrelative information. Together with attention technique which makes the decoding process only focus on the most related source words, the RNN encoder-decoder framework is expected to be able to handle long sequences and consider the globally related information. However, the practical situation is that RNNs tend to forget old history information, especially the far older one. Sometimes the older information is indispensable for generating proper translation, e.g., for the source sentence ``take the heavy box away'', when translating ``away'', ``take'' should be considered together. In addition, it has been proven that using phrases rather than words in SMT \cite{koehn2003statistical} brings performance improvement, while in NMT the attention is only modeled in the unit of words. In the same sense, improvement is expected if attention is operated on more words rather than one.

Moreover, NMT produces the representation for the source by running through the source words sequentially with a bidirectional RNN \cite{schuster1997bidirectional}, so it only employs word order information and ignores the relation between words. Although some researchers have demonstrated that NMT is able to capture certain syntactic phenomena (e.g. subject-verb agreement) without external syntactic information \cite{linzen2016assessing,shi2016does}, there are some other works which has shown their superior performance by modeling word relationship explicitly. However, these works usually need to introduce external syntactic knowledge or connect words according to their relations in the syntactic structure \cite{sennrich2015neural,bastings2017graph,aharoni2017towards,li2017modeling}.

In this paper, we present a method to refine the NMT based on the above two points. The main idea is to learn relationship between the source word pairs. Corresponding to the first point, our method employs Convolutional Neural Networks (CNNs) to collect local information around one word and relates each word with its neighbors, which ensures the subsequent operations are performed in the unit of multiple words. As for the second point, Relation Network (RN) \cite{santoro2017simple} is introduced to establish pairwise relationship between words, meanwhile, there's no need to attain external input of syntactic knowledge. In this way, our model can memorize all words ahead and behind via additional connection between words no matter how distant they are. In the RNs, the representations of the source words produced by RNNs are taken as objects and the relationships between them are reasoned.

Specifically, our method introduces a RN component between the encoder and the attention layer in the RNN encoder-decoder framework \cite{sutskever2014sequence,Bahdanau}. The RN component is composed of three layers: first, the CNN layer slides window along the output of the encoder to capture information among multiple words around one word, then the graph propagation layer constructs a fully connected graph with the information of one window as one node and transfers messages along the edges, so that each node can collect the information from all other nodes, and last the multi-layer perceptron layer transforms the information of each node to the form which is suitable for the attention component to use. We performed experiments on several datasets and got significant improvements over vanilla NMT and SMT systems. Besides, our model significantly outperforms two other models, which introduced latent variables to capture the implicit semantics and employed explicitly external syntactic knowledge respectively.

\section{Background} 

As the main idea of our method is to introduce relation networks into the attention-based NMT \cite{Bahdanau} to learn word relationship and keep all source words in memory, in this section we will briefly describe the baseline model -- the attention-based NMT first and the technique used in this paper -- relation networks.
 
\subsection{Attention-based NMT}

\begin{figure}[!t]
    \centering
    \includegraphics[scale=0.32]{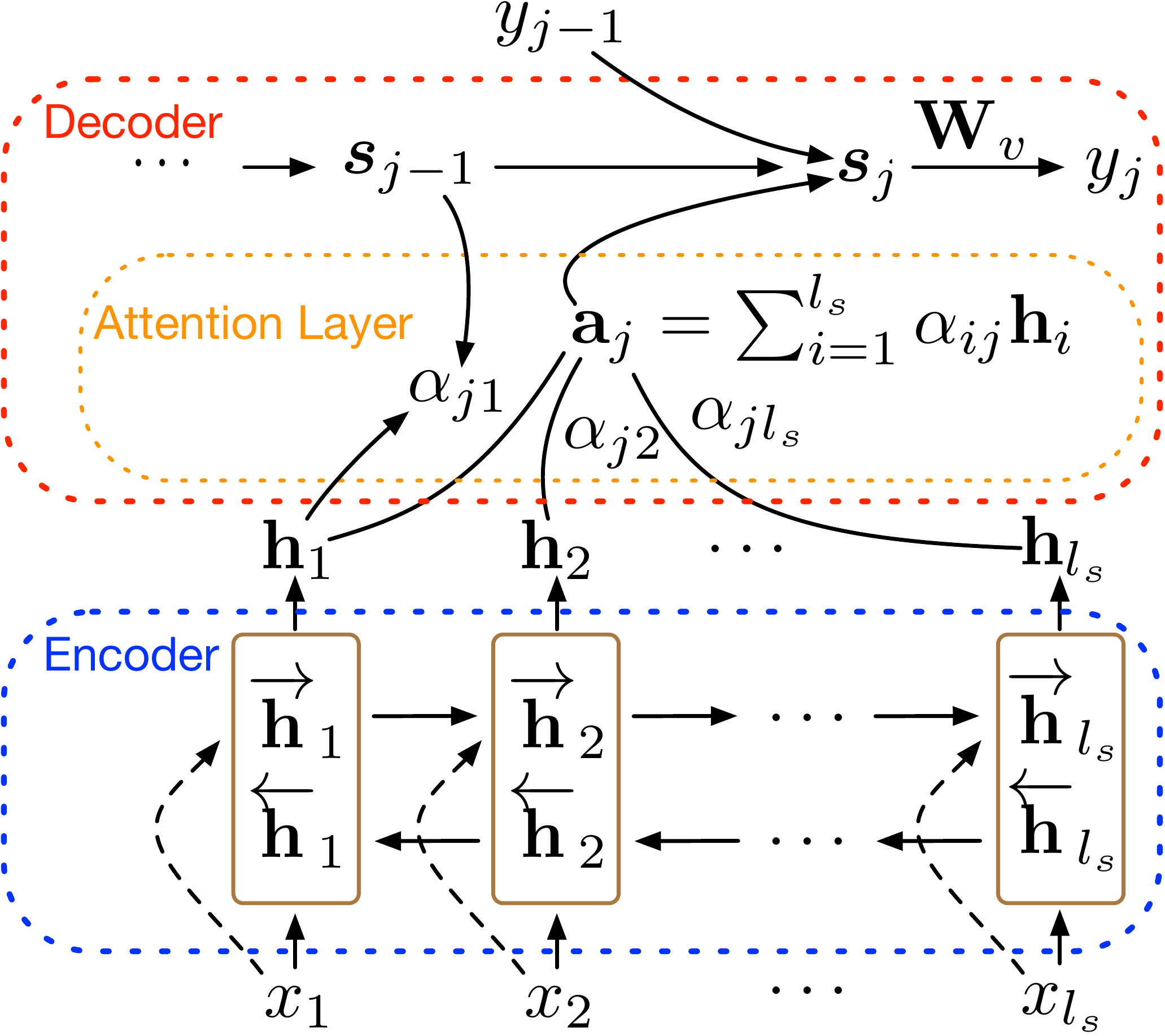}
    \caption{The architecture of attention-based NMT}
    \label{fig:nmt}
\end{figure}

The attention-based NMT follows the encoder-decoder framework, with an additional attention module. It works on the assumption that the source sentence and the target translation share a common continuous space. It first encodes the source sentence into a continuous space and then performs decoding based on this space, meanwhile, employing attention to indicate the relevance of each source word to the current translation. Figure \ref{fig:nmt} shows the architecture of the attention-based NMT \cite{Bahdanau}, which is composed of three components: the encoder, the attention layer and the decoder.

{\bf The Encoder}
The encoder uses a pair of GRUs to run through source words bidirectionally to get two sequences of hidden states, which are concatenated to produce corresponding hidden state for the $i$-th source word
\begin{eqnarray} \label{eq:encoder}
\overrightarrow{\bm{\mathrm{h}}}_i = \overrightarrow{\bm{\mathrm{GRU}}}\left(x_i, \overrightarrow{\bm{\mathrm{h}}}_{i-1}\right); \qquad 
\overleftarrow{\bm{\mathrm{h}}}_i =  \overleftarrow{\bm{\mathrm{GRU}}}\left(x_i, \overleftarrow{\bm{\mathrm{h}}}_{i+1}\right);  \qquad
\bm{\mathrm{h}}_i = \left[{\overrightarrow{\bm{\mathrm{h}}}_i};{\overleftarrow{\bm{\mathrm{h}}}_i}\right]
\end{eqnarray}

{\bf The Attention Layer}
The attention layer aims to extract the source information (called attention) which is highly related to the generation of the current target word. To get the attention of the $j$-th decoding step, the correlation degree between current target word $y_j$ and $\bm{\mathrm{h}}_i$ is first evaluated as
\begin{equation} \label{eq:attend}
    e_{ij}=\bm{\mathrm{v}}_a^T \tanh(\bm{\mathrm{W}}_a\bm{s}_{j-1} + \bm{\mathrm{U}}_a\bm{\mathrm{h}}_i)
\end{equation}
Then, for the $j$-th decoding step, the correlation degree is normalized over the whole source sequence, all source hidden states are added weightedly according to the normalized correlation degree to obtain the attention $\bm{\mathrm{a}}_j$
\begin{equation} \label{eq:alpha}
    \alpha_{ij} = \frac{\exp \left( e_{ij} \right)}{\sum_{i'=1}^{l_s} \exp \left( e_{i'j} \right)}; \qquad
    \bm{\mathrm{a}}_j = \sum\nolimits_{i=1}^{l_s}\alpha_{ij}\bm{\mathrm{h}}_i
\end{equation}

{\bf The Decoder}
The decoder first employs a variant of GRU to roll the target information according to previous target word $y_{j-1}$, previous hidden state $\bm{s}_{j-1}$ and the attention $\bm{\mathrm{a}}_j$. The details are described in Bahdanau et al.~\shortcite{Bahdanau}. The current target hidden state $\bm{s}_j$ is calculated by
\begin{equation} \label{eq:decode:s}
    \bm{s}_j = g(y_{j-1}, \bm{s}_{j-1}, \bm{\mathrm{a}}_j)
\end{equation}
After that, the decoder gives a probability distribution over all the words in the target vocabulary and selects the target word with the highest probability as the output of the current step
\begin{equation} \label{eq:decode:p}
   p(y_j|\bm{\mathrm{y}}_{<j}, \bm{\mathrm{x}}) \propto \exp(f(\bm{s}_j, y_{j-1}, \bm{\mathrm{a}}_j) \cdot \bm{\mathrm{W}_v}) 
\end{equation}
where $f$ stands for a linear transformation and $\bm{\mathrm{W}_v}$ is a weight matrix.

\subsection{Relation Networks} \label{sec:rn}
A relation network (RN) is a neural network with a structure integrated for relational reasoning. The RN is designed to constrain the functional form of a neural network so that it can capture the core common properties of relational reasoning. Hence its capability of computing relations is inherent without needing to be learned specially.

Formally, given a set of input ``objects'' denoted as $\mathrm{O} = \{o_1, o_2, \cdots , o_n\}$, RN can be formed as a composition function of objects \cite{santoro2017simple}, represented as
\begin{equation} \label{eq:rn}
\mathrm{RN(O)}=f_{\phi}\left(\sum\nolimits_{i,j}{g_{\theta}\left(o_i, o_j\right)}\right)
\end{equation}
where $o_i$ is the $i$-th object, and $f_{\phi}$ and $g_{\theta}$ are functions used to calculate relations. Multi-layer perceptrons are often used for $f_{\phi}$ and $g_{\theta}$, as their parameters are learnable synaptic weights, making RNs end-to-end differentiable. Here the role of $g_{\theta}$ is to infer how two objects are related, or whether they are related, and hence the output of $g_{\theta}$ can be treated as ``relations''.

\begin{figure*}[!hbt]
    \centering
    \includegraphics[scale=0.4]{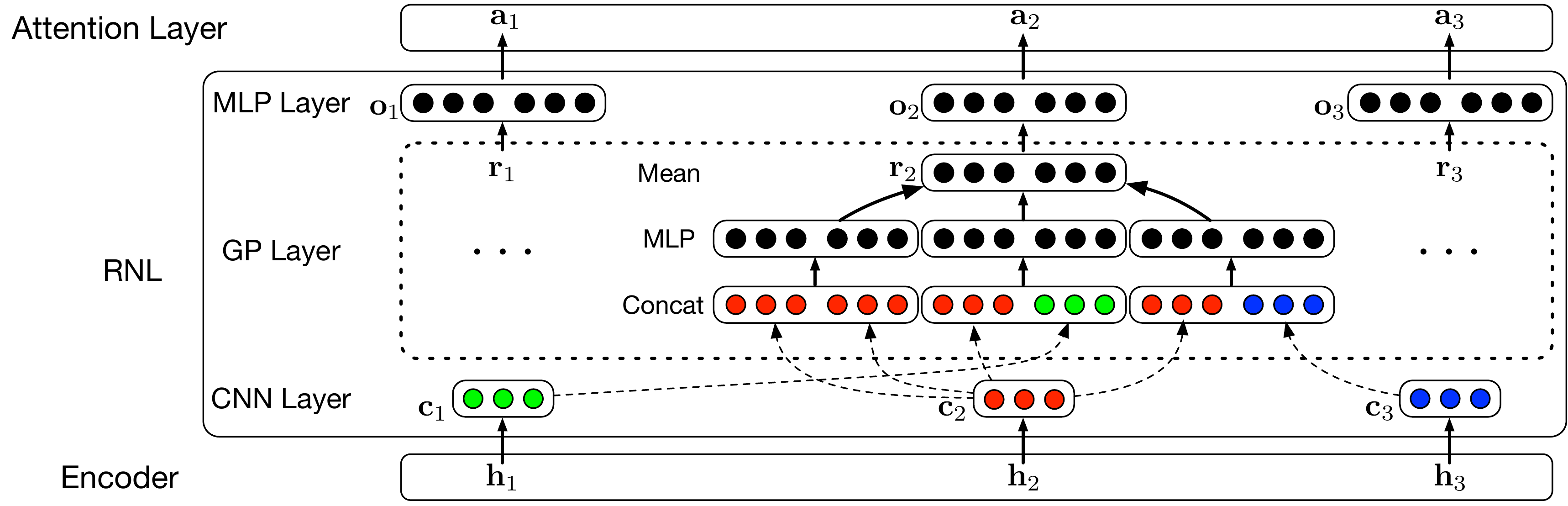}
    \caption{NMT with one RNL. Residual connection is not embodied here. The kernel width of the CNN layer is $3$. We take the second word colored in red as an example to show the operations in the RNL, where three colors of green, red and blue indicate the information from the CNN layer of the first, second and third word, respectively.}
\label{fig:rnNMT}
\end{figure*}

\section{NMT with Relation Networks}

In this paper, we introduce a Relation Network Layer (denoted as RNL) on the basis of the attention-based NMT \cite{Bahdanau} and frame it between the encoder and the attention layer. The RNL first employs CNNs to collect information in the unit of multi-words rather than one single word, then takes the outputs of CNNs as objects and makes them fully connected to build a graph propagation layer and associate with each other, finally transforms the acquired representations with word relations via MLP into the form suitable for the attention layer to use.
Next, the outputs of the RNL are directly fed into the attention layer, so the RNL can still fit the encoder-decoder framework well. The architecture of our RNL is shown in Figure \ref{fig:rnNMT}. Briefly, the RNL is composed of three components: the CNN layer, the Graph Propagation (GP) layer and the Multi-layer Perceptron (MLP) layer.

{\bf The CNN Layer} CNNs are used to collect local information around one word. In this way, not only the information of a single word but their neighbors are considered. The number of neighbors to be considered depends on the kernel width $k$ but can also vary by stacking several convolution layers, e.g., stacking $2$ convolution layers with the kernel width $k=3$ can collect information from $5$ words at the same time.

In the CNN layer, the input is the hidden states produced by the bidirectional GRUs (Bi-GRUs), denoted as $\bm{\mathrm{h}}=\{\bm{\mathrm{h}}_{1}, ..., \bm{\mathrm{h}}_{i}, ..., \bm{\mathrm{h}}_{l_s}\}$, so each source word is represented by its hidden state. A filter is applied to convolute over a window of $k$ words to get the convolutional representation. 
Given the $i$-th source word and its hidden state $\bm{\mathrm{h}}_{i}\in\mathbb{R}^{d}$, the hidden states 
covered by the window with the width of $k$ are concatenated and then are fed to the filter where we denote the concatenated vector as 
$\bm{\mathrm{h}}_{i}^{k} = \left[\bm{\mathrm{h}}_{i-\lfloor{(k-1)/2}\rfloor}; \cdots; \bm{\mathrm{h}}_{i}; \cdots; \bm{\mathrm{h}}_{i+\lfloor{(k-1)/2}\rfloor}\right]$. For the first and last $\lfloor{(k-1)/2}\rfloor$ words of a sentence, the hidden state $\bm{\mathrm{h}}_{i}$ with $i<1$ or $i>l_s$ are set to zeros (padding).
Then the filtering process mentioned above can be formed as

\begin{equation} \label{eq:rn:c}
\bm{\mathrm{c}}_{i} = f\left(\bm{\mathrm{W}}_{cnn}\bm{\mathrm{h}}_{i}^{k} + \bm{\mathrm{b}}_{cnn}\right)
\end{equation}

\noindent $\bm{\mathrm{W}}_{cnn}\in\mathbb{R}^{k\times{d}}$ is the convolution weights and $\bm{\mathrm{b}}_{cnn}$ is the bias, where the two together define a linear operation.  $f$ is the leaky RELU with the coefficient $0.1$ to control the angle of the negative slope. In the RNL, leaky RELU is used as all the nonlinear activation functions.
The output of the CNN layer is $\bm{\mathrm{c}}=\{\bm{\mathrm{c}}_{1}, ..., \bm{\mathrm{c}}_{i}, ..., \bm{\mathrm{c}}_{l_s}\}$.

{\bf The GP Layer}
The GP layer is used to learn the relationships between source words.
It adopts the outputs of the CNN layer $\bm{\mathrm{c}}=\lbrace\bm{\mathrm{c}}_{1}, \bm{\mathrm{c}}_{2}, ..., \bm{\mathrm{c}}_{l_s}\rbrace$ as input and formulates the relationships between them into a graph. 
Here $\bm{\mathrm{c}_{i}}$ can be thought as the object mentioned in Section \ref{sec:rn}. 
In this graph, each input $\bm{\mathrm{c}_{i}}$ is taken as a node and has edges connected to all other nodes. Then information flows along the edges and each node receives messages from all its direct neighbors. We call this process graph propagation. 

After graph propagation process, another sequence of vectors $\lbrace\bm{\mathrm{r}}_{1}, \bm{\mathrm{r}}_{2}, ..., \bm{\mathrm{r}}_{l_s}\rbrace$ is produced.
The generation of $\bm{\mathrm{r}}_{i}$ can be decomposed into three steps:

\begin{itemize}
\item Each input vector $\bm{\mathrm{c}}_{i}$ in $\bm{\mathrm{c}}$ is concatenated with all vectors in $\bm{\mathrm{c}}$ (including itself) to get a set of vectors $\bm{\mathrm{C}}_i=\lbrace\bm{\mathrm{c}}_{i1}, \cdots \bm{\mathrm{c}}_{ij}, \cdots, \bm{\mathrm{c}}_{il_s}\rbrace$ where
$\bm{\mathrm{c}}_{ij} = \left[\bm{\mathrm{c}}_i;\bm{\mathrm{c}}_j\right]$.

\item Each $\bm{\mathrm{c}}_{ij}$ is converted into vector $\bm{\mathrm{r}}_{ij}$ by a $4$-hidden-layers MLP. The conversion with $1$-hidden-layer MLP can be represented as
\begin{equation} \label{eq:rn:r_ij}
\bm{\mathrm{r}}_{ij} = f\left(\bm{\mathrm{W}}_{gp}  \bm{\mathrm{c}}_{ij} + \bm{\mathrm{b}}_{gp}\right)
\end{equation}
\item Average over all the outputs above to get the final representation for the $i$-th source word
\begin{equation} \label{eq:rn:r}
    \bm{\mathrm{r}}_{i} = \frac{1}{l_s}\sum\nolimits_{j=1}^{l_s}\bm{\mathrm{r}}_{ij}
\end{equation}
\end{itemize}

{\bf The MLP Layer}
There are several nonlinear transformations which map the inputs into different vector spaces in the GP layer. In order to reduce computation complexity, the output features size of the nonlinear transformations is set to small. Hence we use another MLP layer to map the feature back into the original space, usually the same as that of $\bm{\mathrm{h}}_i$ to have more powerful representation. The final state $\bm{\mathrm{o}}_i$ for the $i$-th source word after the entire RN layer can be got by another $2$-hidden-layers MLP, $1$-hidden-layer MLP can be written as
\begin{equation} \label{eq:rn:o}
\bm{\mathrm{o}}_i = f(\bm{\mathrm{W}}_{mlp} \bm{\mathrm{r}}_i + \bm{\mathrm{b}}_{mlp})
\end{equation}

{\bf Residual}
Stacking technique is used in our method. Concretely, we stack multiple layers inside the encoder and meanwhile apply residual connection for two adjacent layers. Assume $\mathrm{h}_{in}^{l}$ and $\mathrm{h}_{out}^{l}$ are the input and the output of the $l$-th layer, respectively, then residual connection is conducted to get the final output of the $l$-th layer in the following two steps. First, the input and the output of the $l$-th layer are added together:
\begin{equation}
\mathrm{h}^{l} = \mathrm{h}_{in}^{l} + \mathrm{h}_{out}^{l}
\end{equation}

Next, dense concatenation \cite{huang2017densely} is employed to receives features from all previous layers and the final output of the $l$-th layer is produced by
\begin{equation}
\mathrm{h}_{dc}^{l} = \bm{\mathrm{W}}_{dc} \left[\mathrm{h}^{1};\mathrm{h}^{2};\cdots;\mathrm{h}^{l}\right] + \bm{\mathrm{b}}_{dc}
\end{equation}

\noindent where weight matrix $\bm{\mathrm{W}}_{dc}$ and bias $\bm{\mathrm{b}}_{dc}$ are adjusted to map the dense-concatenated vectors into the same feature space as the input. Then $\mathrm{h}_{dc}^{l}$ is fed to the next layer which means $\mathrm{h}_{in}^{l+1}=\mathrm{h}_{dc}^{l}$.

\section{Related Work}

Many researchers have worked on learning the relationships of the source words to improve translation performance. One line is to refine source presentations by adding relationships between source words or between source and target words, with the main architecture remaining the RNN encoder-decoder framework.
Sennrich et al.~\shortcite{sennrich2016linguistic} enriched source representations with POS tags, dependency labels and other linguistic features.
Bastings et al.~\shortcite{bastings2017graph} employed graph convolutional networks to model relations of words in dependency trees for the source embeddings to include these relations. These two models both require extra supervised syntax input while our method does not need external knowledge and learn the relationship by its own.

Another line is to change the structure of the neural network. Gehring et al.~\shortcite{gehring2016convolutional} and Gehring et al.~\shortcite{gehring2017convolutional} proposed to substitute the conventional RNN encoder with the CNN encoder in order to train faster. They employed stacked CNNs to capture relationships between source words which can be calculated simultaneously, not like RNNs, the computation of which is constrained by temporal dependencies. The attention scores are also computed based on the output of the CNNs and the decoder is still the RNN-based decoder. Vaswani et al.~\shortcite{vaswani2017attention} is another work to eschew the recurrence. It instead relied entirely on the attention mechanism to draw the global dependencies between input and output. Su et al.~\shortcite{su2018variational} introduced latent random variables into the decoder of NMT and generated these variables recurrently to capture the global semantic contexts and model strong and complex dependencies among target words at different timesteps.

Our method still follows the RNN encoder-decoder framework, giving full play to the advantages of RNNs, which transfers information through words bidirectionally. In addition, we also employs RNs in our method to connect the source words explicitly, further captures relationships between source words without any external knowledge injection, which enables the model to learn the relationships itself and facilitates easy application.

\section{Experiments}

In the experiment section, we first compare our system with two baseline systems on a Chinese-English (Zh-En) dataset and the WMT17 English-German (En-De) dataset, then compare our method with a related approach on the WMT16 En-De dataset. Finally, we give some analyses about our method in different aspects.

\subsection{Data Preparation}

We performed experiments on three datasets:

\textbf{NIST} The training data consisted of $1.25$M Zh-En parallel sentence pairs with $25$M Chinese tokens and $27$M English tokens\footnote{We chose LDC2002E18, LDC2003E07, LDC2003E14, Hansard’s portion of LDC2004T07, LDC2004T08 and LDC2005T06 from the LDC corpora. There were $1.11$M sentence pairs left after filtering.}. We used NIST 2002 test dataset ($878$ sentences) as the validation set, and another four NIST test datasets as the test datasets: NIST 2003 (MT03), NIST 2004 (MT04), NIST 2005 (MT05) and NIST 2006 (MT06), which contain $919$, $1788$, $1082$ and $1357$ sentences respectively.

\textbf{WMT17} The training data was composed of $5.6$M En-De preprocessed parallel sentence pairs \footnote{\url{http://data.statmt.org/wmt17/translation-task/preprocessed}} with $141$M English tokens and $194$M German tokens. The test dataset of newstest2014 ($3003$ sentences) was used as the validation set and the following test datasets were used as the test datasets: newstest2015 ($2169$ sentences), newstest2016 ($2999$ sentences) and newstest2017 ($3004$ sentences). Besides, $8k$ merging operations were performed to learn byte-pair encodings (BPE) \cite{sennrich2015neural} on the target side of the parallel training data.

\textbf{WMT16} We conducted experiments on WMT16 dataset, the same dataset as the work of Bastings et al.~\shortcite{bastings2017graph} for comparison. We kept the same settings as those in Bastings et al.~\shortcite{bastings2017graph}:
The original dataset consists of $4500966$ sentence pairs, with $4173550$ left after filtering pairs which contains more than $50$ tokens on either side after tokenization. newstest2015 and newstest2016 were used as the validation set and test dataset, respectively. $16k$ BPE merging operations were conducted on the target side of the bilingual training data.

For WMT16 dataset, case-sensitive $4$-gram BLEU score \cite{papineni2002bleu} was reported by using the {\em multi-bleu.pl} script. The results on the other two datasets were evaluated with case-insensitive $4$-gram BLEU score.

\subsection{Systems}

Results of five systems on different datasets were reported:

\textbf{RNNsearch}
We implemented the attention-based NMT of Bahdanau et al.~\shortcite{Bahdanau} by PyTorch framework\footnote{\url{http://pytorch.org}} with the following settings: the length of the sentences on both sides was limited up to $50$ tokens with $30$K vocabulary, and the source and target word embedding sizes were both set to $512$, the size of all hidden units in both encoder and decoder RNNs was also set to $512$, and all parameters were initialized by using uniform distribution over $\left[-0.1,0.1\right]$. The mini-batch stochastic gradient descent (SGD) algorithm was employed to train the model with batch size of $80$. In addition, the learning rate was adjusted by Adadelta optimizer \cite{zeiler2012adadelta} with $\rho=0.95$ and $\epsilon=1e\textnormal{-}6$. Dropout was applied on the output layer with dropout rate of $0.5$. The beam size was set to $10$.

\textbf{RNNsearch$^{\star}$}
This system is an improved version of RNNsearch where the decoder employs a conditional GRU layer with attention module, consisting of two GRUs and an attention module for each step\footnote{\url{https://github.com/nyu-dl/dl4mt-tutorial/blob/master/docs/cgru.pdf}}. Specifically, Equation \ref{eq:decode:s} is substituted with the following two equations:
\begin{equation} \label{imp_dec}
    \tilde{\bm{s}}_j = \bm{\mathrm{GRU}}_1(y_{j-1}, \bm{s}_{j-1}); \qquad 
    \bm{s}_j = \bm{\mathrm{GRU}}_2(\bm{\mathrm{a}}_j, \tilde{\bm{s}}_j)
\end{equation}
Besides, for the calculation of attention in Equation \ref{eq:attend}, $\bm{s}_{j-1}$ is replaced with $\tilde{\bm{s}}_{j-1}$. The other components of the system keep the same as RNNsearch. We used the same settings for RNNsearch and RNNsearch$^{\star}$.

\textbf{VRNMT}
A novel Variational Recurrent NMT (VRNMT) model, proposed by Su et al.~\shortcite{su2018variational}, captures more semantic context and complex dependencies among target words by generating latent random variables recurrently in the NMT decoder.

\textbf{BiRNN+GCN}
This is the model presented by Bastings et al.~\shortcite{bastings2017graph}. They incorporated dependency syntactic structure into the bidirectional RNN (BiRNN) encoder of NMT and modeled the relation among the source words by using graph convolutional networks (GCNs).

\textbf{\textsc{Rnmt}}
Our system was implemented by embedding the RNLs into the Bi-GRUs of RNNsearch$^{\star}$. The overall structure used alternatively stacked GRUs and RNLs, in which the two GRU layers are in opposite direction. Inside the RNL, the GP layer employed a $4$-hidden-layers MLP (shown in Equation \ref{eq:rn:r_ij}) and the MLP layer contained $2$ hidden layers (as in Equation \ref{eq:rn:o}). For the Zh-En translation task, two convolution layers with kernel width of $1$ and $3$ were stacked, the output channel sizes of CNN were $128$ and $256$ respectively, followed by batch normalization (BN) \cite{ioffe2015batch} with learnable parameters, and MLP contained $256$ units. For the En-De translation task, only one convolution layer was used with kernel width of $3$, the output channel size was $96$, $128$ was adopted as the hidden size of MLP. All of the other settings were the same with those of RNNsearch$^{\star}$.

\begin{table}
\centering
\renewcommand\arraystretch{1.2}
\begin{tabular}{l||c|c|c|c|c}
{\bf Systems }  & {\bf MT03} & {\bf MT04} & {\bf MT05} & {\bf MT06} & {\bf Average} \\ \hline
RNNsearch    & $33.70$ & $36.15$ & $31.81$ & $32.71$ & $33.59$ \\ \hline
RNNsearch$^{\star}$   & $37.93$ & $40.53$ & $36.65$ & $35.80$ & $37.73$ \\ \hline
VRNMT        & $38.08$ & $41.07$ & $36.82$ & $36.72$ & $38.17$ \\ \hline
\textsc{Rnmt}  & {\bf 39.24$^{\ast}$} & {\bf 42.01$^{\ast}$} & {\bf 37.79$^{\ast}$} & {\bf 37.81$^{\ast}$} & {\bf 39.21} \\
\end{tabular}
\caption{Performance comparison on NIST datasets. $\ast$ is used to indicate the improvement over RNNsearch$^{\star}$ is statistically significant \cite{collins2005clause} ($p<0.01$). }
\label{zhen_systems_compare}
\end{table}

\begin{table*}
\centering
\begin{minipage}{0.45\linewidth}
\centering
\renewcommand\arraystretch{1.2}
	\begin{tabular}{l||c|c|c|c}
		{\bf Systems }  & {\bf test15} & {\bf test16} & {\bf test17} & {\bf Avg.} \\ \hline
		RNNsearch    & $17.3$ & $20.9$ & $16.6$ & $18.3$ \\ \hline
		RNNsearch$^{\star}$  & $21.4$ & $25.6$ & $20.1$ & $22.4$ \\ \hline
		\textsc{Rnmt}  & {\bf 22.7$^{\ast}$} & {\bf 27.8$^{\ast}$} & {\bf 21.8$^{\ast}$} & {\bf24.1} \\
	\end{tabular}
	\caption{Performance comparison on WMT17 En-De datasets.}
    \label{ende_systems_compare}
\end{minipage}
\qquad\quad
\begin{minipage}{0.45\linewidth}
\centering
\renewcommand\arraystretch{1.2}
	\begin{tabular}{l||c}
		{\bf Systems}	&	{\bf test16} \\ \hline
		BiRNN+GCN       & $23.9$ \\ \hline
   		\textsc{Rnmt}   & {\bf 25.4}
  	\end{tabular}
  	\caption{Performance comparison with the related work on the WMT16 En-De dataset.}
  	\label{ende_related_work_comparing}
\end{minipage}
\end{table*}

\subsection{Performance Comparison}

We compared our system \textsc{Rnmt} with the two baseline systems RNNsearch and RNNsearch$^{\star}$ both on the NIST Zh-En and the WMT17 En-De translation tasks. As \textsc{Rnmt} was implemented on the basis of RNNsearch$^{\star}$, in the strict sense, RNNsearch$^{\star}$ is the baseline.
From the results shown in Table \ref{zhen_systems_compare}, we can see that $\textsc{Rnmt}$ significantly improves translation quality on all test datasets and outperforms RNNsearch$^{\star}$ by $1.48$ BLEU points averagely on the Zh-En dataset. Besides, comparison between our model to VRNMT shows that proposed simple model stably produces better performance on all test datasets and outperforms VRNMT $1.04$ BLEU score on average.

\begin{figure}[!ht]
\centering
\noindent
\includegraphics[scale=0.23]{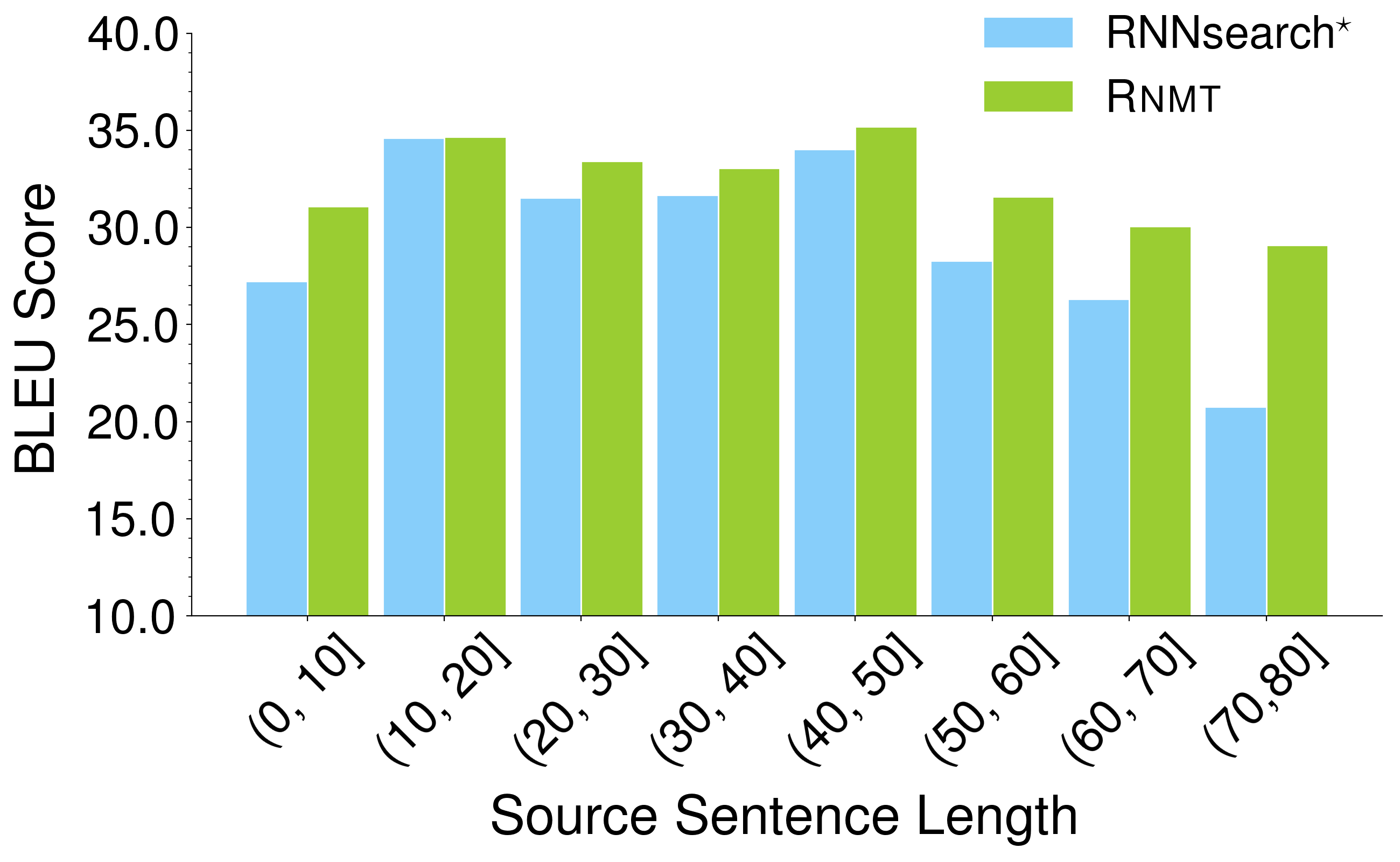}
\caption{Results on different bins contain sentences of length within corresponding spans.}
\label{fig:length}
\end{figure}

On the WMT17 En-De dataset, as shown in Table \ref{ende_systems_compare}, $\textsc{Rnmt}$ shows superiority on three test datasets stably, and averagely achieves the gains of $1.7$ BLEU points over RNNsearch$^{\star}$, with only $4.1$M parameters more. Given the above results, we can conclude that RN can indeed learn the relationships between the source words and these relationships are useful and bring improvement on the translation performance.

We also compared our method with the work of Bastings et al.~\shortcite{bastings2017graph} which requires the injection of external syntactic knowledge, to see whether the relationships produced by RNs can lead to better translation than the syntax from supervised learning. The results in Table \ref{ende_related_work_comparing} show that our system can achieve an improvement of $1.5$ BLEU scores.
We believe that the relationships of the source words derived from RNs do not necessarily conform to human cognition, but it can be simultaneously tailored with the other parts of the translation system. In this way, RNs can generate the relationships more suitable for the NMT.

\subsection{Impact of Input Length}

One motivation of adding RNs is that RNNs tend to forget the distant history which RNs memorize it by explicitly introducing relations between pairs of words. Therefore, we assume that our method suppose to bring greater improvement on relative long sentences, which contains more distant history information than shorter ones that usually forgotten by RNNs. Based on this sense, we split the source sentences in the MT03 test dataset into different bins according to their length and evaluated BLEU scores of the translations from RNNsearch$^{\star}$ and $\textsc{Rnmt}$ on the different bins, respectively. 

The results are shown in Figure \ref{fig:length}. In the bins holding sentences no longer than $50$, the BLEU scores of the two systems are close to each other. When the sentence length surpasses $50$, $\textsc{Rnmt}$ shows its superiority over RNNsearch$^{\star}$. As the sentence length grows, the difference becomes increasingly large. This verifies the deduce that our method can not only memorize history information but capture the relationship between words, both of which are beneficial to translate long sentences.

\subsection{Word Alignment}

\begin{table}[h!]
\centering
\renewcommand\arraystretch{1.1}
\begin{tabular}{l||c|c}
{\bf Systems }       & {\bf BLEU}  & {\bf AER} \\ \hline
RNNsearch$^{\star}$  & 22.40       & 46.76 \\ \hline
\textsc{Rnmt}   	 & {\bf 24.12} & {\bf 45.66} \\
\end{tabular}
\caption{Comparison of alignment quality on NIST Zh-En translation task.}
\label{tb:align}
\end{table}

In this section, we will verify the translation performance of our model from another perspective. Intuitively, the better translation should have better alignment to the source sentence, so we evaluated the quality of the alignments derived from the attention module of the NMT using Alignment Error Rate (AER) \cite{och2003minimum}. We did this experiment on the artificially aligned dataset from Liu and Sun~\shortcite{liu2015contrastive} which contains $900$ Zh-En sentence pairs. The alignments were got in this way for both RNNsearch$^{\star}$ system and our system. When one target word was generated, we retained the alignment link with the highest probability $\alpha_{ij}$ in Equation \ref{eq:alpha}.

The comparison results are shown in Table \ref{tb:align}. It illustrates that our system \textsc{Rnmt} can produce better translations than the baseline RNNsearch$^{\star}$, a difference of $1.72$ BLEU points. Besides, the AER score is $1.1$ points lower than the baseline model. Note that the smaller the AER score, the better the alignment quality.

\begin{figure*}[bt]
\centering
\subfigure[RNNsearch$^{\star}$]{
\includegraphics[scale=0.26]{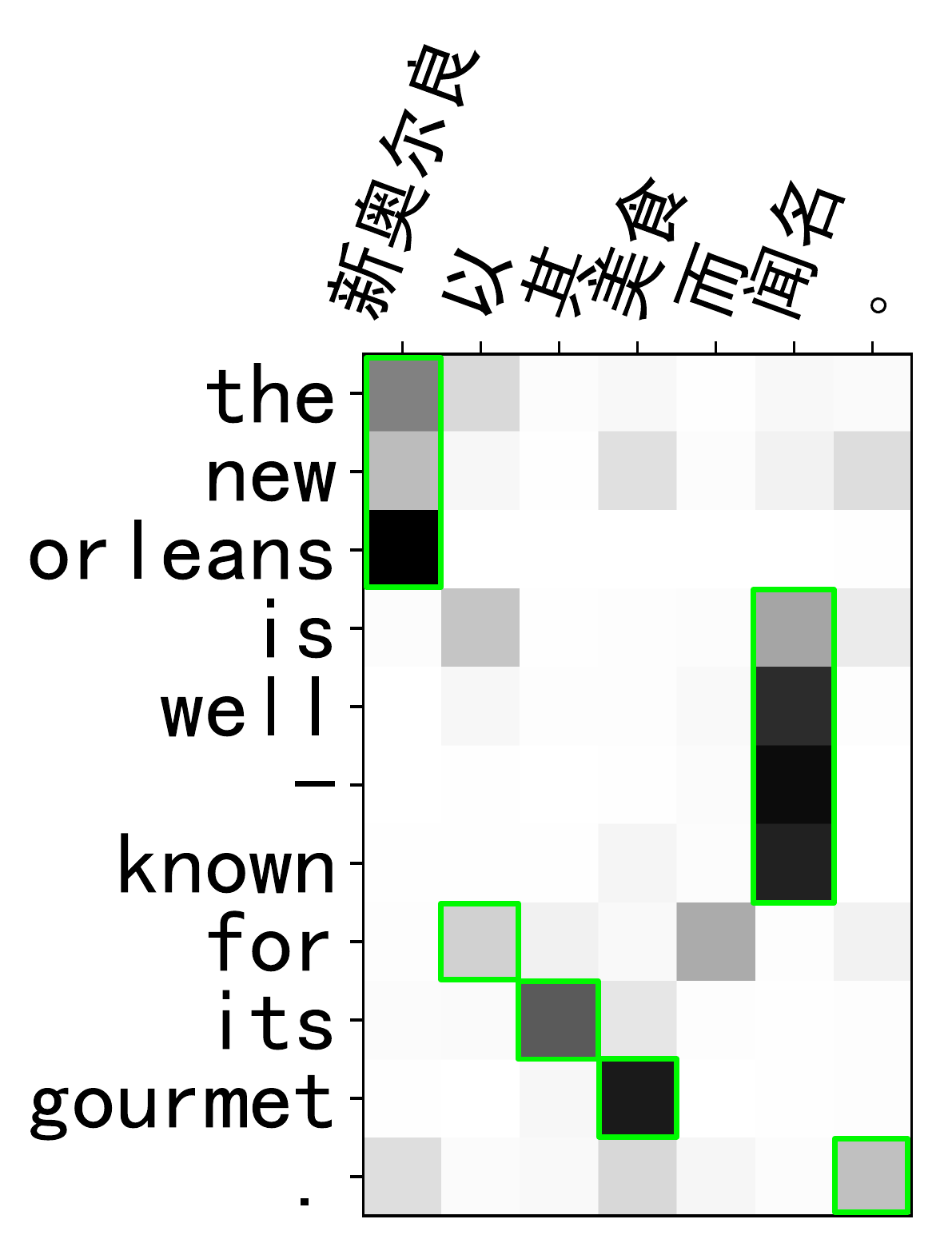}
\includegraphics[scale=0.26]{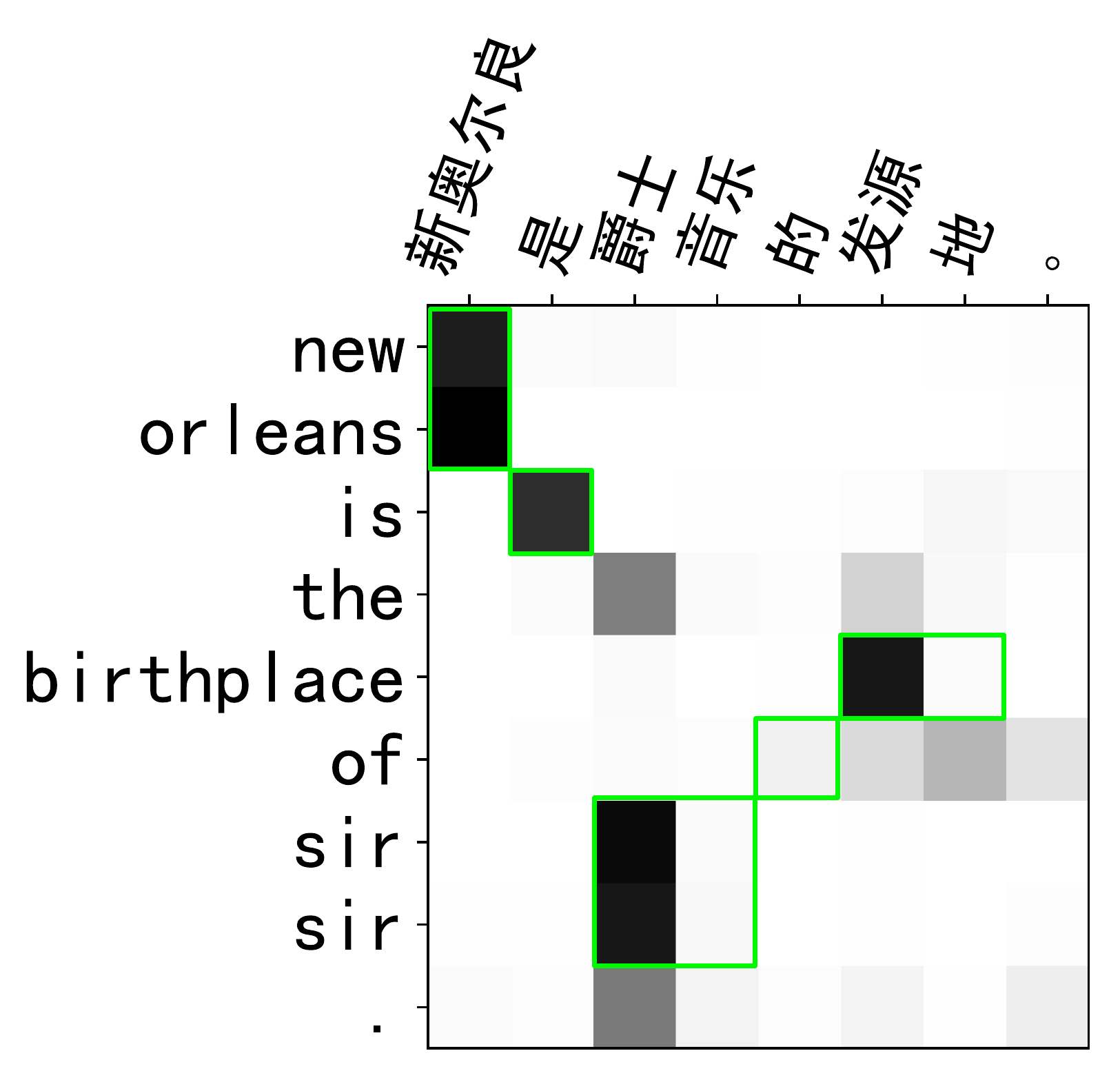}
}
\subfigure[\textsc{Rnmt}]{
\includegraphics[scale=0.26]{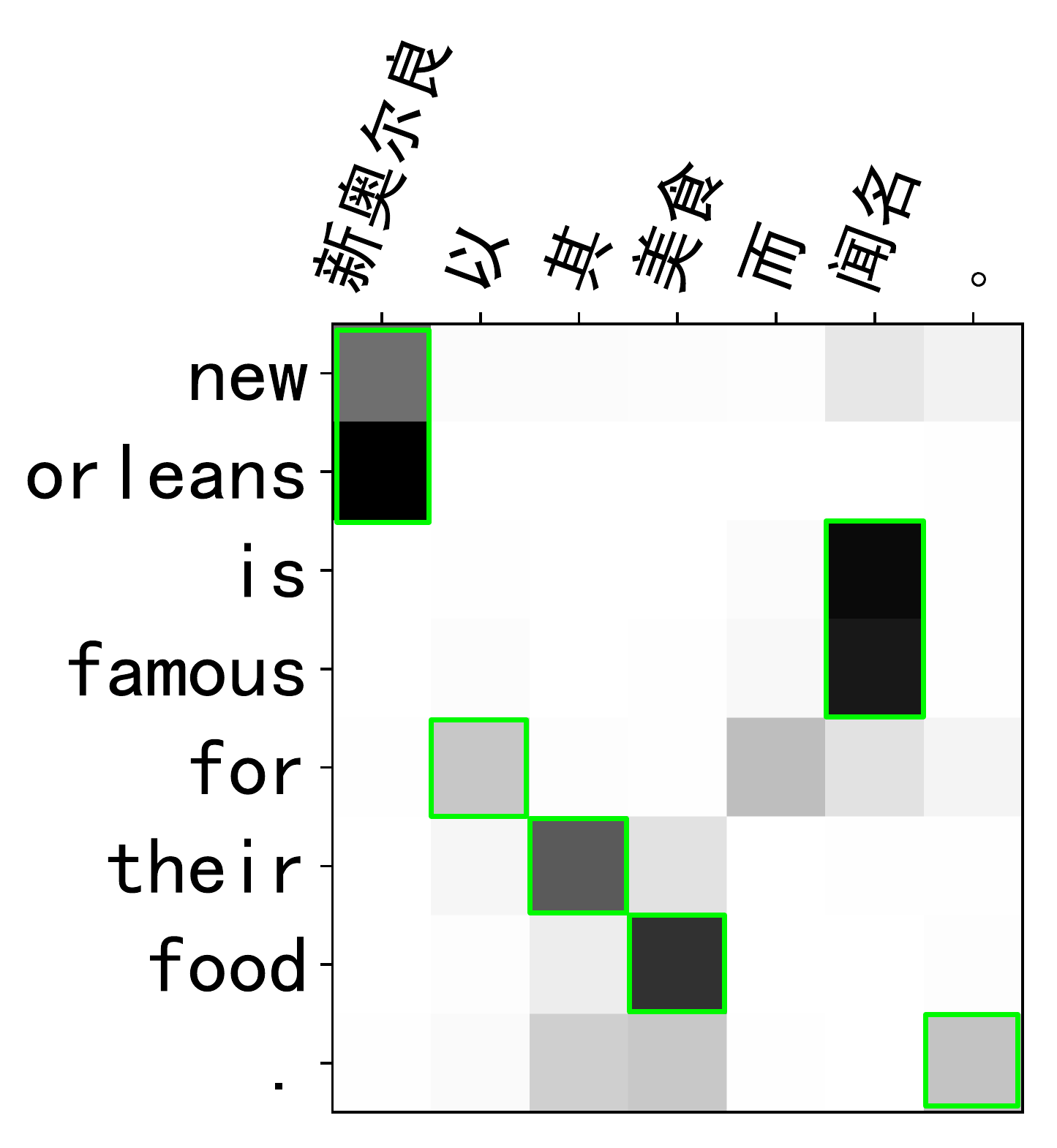}
\includegraphics[scale=0.26]{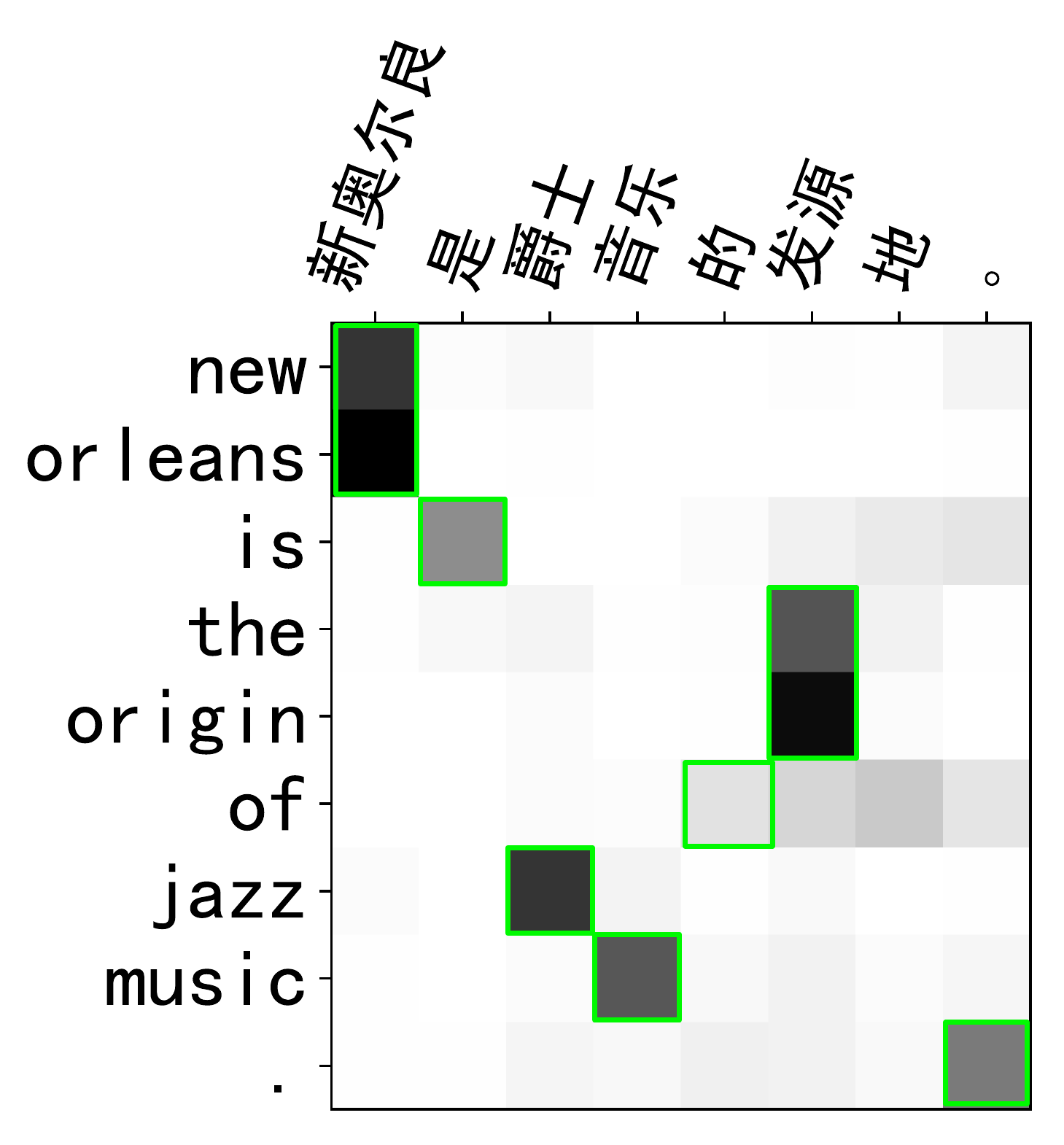}
}
\caption{Word alignment comparison. The green boxes show the manual golden alignments.}
\label{fig:wordalignment}
\end{figure*}

Along with the translation results, we also produce the word alignment matrix based on each target word's attention probability distribution over the whole source sentence. Two source sentences are randomly sampled from websites, both comparisons between baseline alignment and improved alignment generated by RNNsearch$^{\star}$ and \textsc{Rnmt} are shown in Figure \ref{fig:wordalignment}.

For the first example, from the view of source side, it is obviously unreasonable that the Chinese word \textit{yi} is contributed to generate three discontinuous English words \textit{the}, \textit{is} and \textit{for}, grammatical knowledge show that the word \textit{yi} should be only aligned to the English word \textit{for}, just like the result of our model. Besides, on the target's ground, if one Chinese word is translated into an English phrase, all words in the phrase should be aligned to the Chinese word, RNNsearch$^{\star}$ model improperly aligns \textit{new} and \textit{is} to some other irrelevant words besides the correct one. When generating word \textit{is}, almost the whole source sentence should be considered, our model gets more centralized alignment for it.

In the second case, unlike the baseline model, our model produces correct translation \textit{jazz music} for \textit{jueshi yinyue} and alignment. \textit{the} together with \textit{origin} is aligned to the source word \textit{fayuan}, while RNNsearch$^{\star}$ mistakenly aligns \textit{the} to two source words almost with equal probability.

\begin{table*}[!htb]
\centering
\renewcommand\arraystretch{1.2}
{\footnotesize
\resizebox{\textwidth}{!}{
  \begin{tabular}{l|l}
    \toprule
    Source     & \multicolumn{1}{p{14cm}}{我们\ 近年\ 一直\ 倡导\ “\ 诚信\ ”\ ,\ 要\ “\ 打造\ 阳光\ 政府\ ”\ ,\ 要\ 尊重\ 公众\ 的\ “\ 知情权\ ”\ ,\ 要\ 提高\ 行政\ “\ 透明度\ ”\ ,\ 然而\ ,\ 事实\ 距离\ 理想\ 还\ 有\ 很\ 大\ 差距\ 。}  \\
    Reference  & \multicolumn{1}{p{14cm}}{in recent years, we have been advocating "integrity" and we want to "forge a government-in-sunshine", improve the "transparency" of government administration, and respect the public\textquotesingle s "right to know". however, the reality is still very far from ideal.}  \\ 
    RNNsearch$^{\star}$   & \multicolumn{1}{p{14cm}}{in recent years , we have always advocated " honesty " and " build a sunshine government , " and we must respect the public \textquotesingle s " right to understand " and to enhance the " transparency " of the " transparency " of the public .} \\
    \textsc{Rnmt}  & \multicolumn{1}{p{14cm}}{in recent years , we have advocated " integrity " and " build up the sun . " we should respect public " right to know " and improve the \textbf{" transparency " of the public . however , there is still a big gap between reality and ideals .}} \\
    \midrule
    Source     & \multicolumn{1}{p{14cm}}{经过\ 国际\ 奥委会\ 的\ 不懈\ 努力\ ,\ 意大利\ 方面\ 在\ 冬奥会\ 开幕\ 前\ 四\ 天\ 作出\ 让步\ ,\ 承诺\ 冬奥会\ 期间\ 警方\ 不\ 会\ 进入\ 奥运村\ 搜查\ 运动员\ 驻地\ ,\ 但是\ ,\ 药检\ 呈\ 阳性\ 的\ 运动员\ 仍将\ 接受\ 意大利\ 检察\ 机关\ 的\ 调查\ 。}  \\
    Reference  & \multicolumn{1}{p{14cm}}{through the untiring efforts of the ioc, the italian side made concession four days before the winter olympics opened, promising that police would not enter the olympic village to raid athletes\textquotesingle\ quarters during the winter olympics, but athletes tested positive for drugs are still subject to investigations of italian prosecutors.}  \\ 
    RNNsearch$^{\star}$    & \multicolumn{1}{p{14cm}}{through the unremitting efforts of the ioc , the italian side made a concession four days prior to the opening of the international olympic committee .} \\
    \textsc{Rnmt}    & \multicolumn{1}{p{14cm}}{with the unremitting efforts of the international olympic committee , the italian side made a concession in four days before the opening of the $\langle\mathrm{UNK}\rangle$ \textbf{and promised that the police would not be able to search for the athlete \textquotesingle s place during the opening period .}} \\
    \bottomrule
  \end{tabular}
  }
}
  \caption{Translation examples.}
  \label{t3}
\end{table*}

\subsection{Translation Examples}

As shown in Table \ref{t3}, we give two example translations generated from baseline model and proposed model. Comparing the translation results between two systems, we can observe that RNNsearch$^{\star}$ often miss some information of the source sentence, especially for the long sentence. Both of the sentences are complex sentences with long dependent adversative relation, for the first example, the baseline model forgets the information of the long distance clause about \emph{women jinnian yizhi $\cdots$ toumingdu} and ignores to translate the second clause. It similarly happens that, when producing the target text for the second sample, RNNsearch$^{\star}$ loses the information after \emph{chengnuo dongaohui} and fails to capture the latter clause with adversative relation. In addition, another phenomenon observed is that the longer the source sentence is, it is easier to ignore important information for RNNsearch$^{\star}$. However, as can be seen from the boldfaced sections marked in results generated with \textsc{Rnmt}, proposed model with CNN could captures more source information successfully.

Specifically, RNNs are skilled in modeling the order information of a sequence, while CNNs mainly focus on local features around some specific word. Both of them are weak to capture the long-distance dependency information, However, facts prove that proposed relation layer succeeds in alleviating the deficiencies of the two by integrating CNNs with bidirectional RNNs subtly.

\section{Conclusion}

As RNNs are not good at remembering the old history and cannot consider word relationship either, sometimes conventional NMT cannot get enough source information and hence emphasizes too much on the fluency of the target. As a result, it suffers from meaning-drift and generates ``inaccurate'' translation. Even so, NMT can still benefit from the recurrence of RNNs. In this paper, we propose to incorporate RNLs into the attentional NMT. The RNs employs CNNs to collect information around one word and explicitly connect each word with all the other words. In this way, it provides the opportunities for NMT to capture relationship between source words and hence leads to a better source representation.
Our method can get better translation on the NIST Zh-En dataset and the WMT En-De dataset and can even outperform the system with supervised syntactic knowledge.

\section*{Acknowledgements}
We highly appreciate the anonymous reviewers for their precious comments. This work was supported in part by National Natural Science Foundation of China (Nos. 61472428 and 61662077).


\end{CJK*}

\end{document}